# Heuristic Search as Evidential Reasoning*

Othar Hansson and Andrew Mayer

Computer Science Division
University of California
Berkeley, CA 94720

**Abstract**

BPS, the Bayesian Problem Solver, applies probabilistic inference and decision-theoretic control to flexible, resource-constrained problem-solving. This paper focuses on the Bayesian inference mechanism in BPS, and contrasts it with those of traditional heuristic search techniques.

By performing sound inference, BPS can outperform traditional techniques with significantly less computational effort. Empirical tests on the Eight Puzzle show that after only a few hundred node expansions, BPS makes better decisions than does the best existing algorithm after several million node expansions.

## 1 Problem-Solving and Search

Problem-solving may be formalized within the problem-space representation of [17], in which problems are specified by a set of possible *states* of the world and a set of *operators*, or transitions between states. A problem instance is specified by a single *initial state*, $I$, and set of *goal states*, $G$. The states and operators form the nodes and directed arcs of a *state-space graph*. Problem-solving in the state-space requires applying a sequence of operators, or *solution-path*, to state $I$, yielding a state in $G$.

### 1.1 Complete Decision-Trees

Existing state-space search algorithms are adaptations of the game-theoretic techniques for evaluating *decision-trees*. From the perspective of a problem-solver, the state-space graph can be viewed as a decision-tree, whose *root node* is the initial state. In this decision-tree, paths correspond to sequences of operators, and leaves to terminal states of the problem.

These leaves have associated payoffs, or *outcomes*. In the two-player game of chess, for example, the outcomes are win, loss and draw. Similarly, in single-agent path-planning domains, the outcomes describe the desirability of goal states and the paths used to reach them. In theory, one can directly *label* each leaf with its outcome, and recursively label each internal node by assigning it the outcome of its most preferred child. Once the entire tree has been so labelled, the problem-solver need only move from the root node to the neighboring node labelled with the most preferred outcome. The MaxMin procedure[26] is a classic example of this constraint-satisfaction labelling technique.

### 1.2 Traditional Approaches

The complexity of state-space problem-solving arises from the fact that most interesting problems have enormous, densely-connected state-space graphs. Because of resource (e.g., computational) constraints, most problem-solvers will be unable to explore the entirety of the decision-tree before being forced to commit to an action. Rather, one will see only a relatively small portion of the entire decision-tree, and must select an *operator* to apply before knowing the outcomes of all adjacent states with certainty.

How should a *partial* decision-tree be inter-preted? The conventional wisdom in both artificial intelligence and decision analysis is to evaluate the partial decision-tree as a decision-tree proper. Examples of this approach include the "averaging out and folding back" technique in decision analysis[20], and the Minimax algorithm in artificial intelligence[25], both of which trace their origin to MaxMin. In these techniques, *heuristics* are used to estimate the (unknown) outcomes of the *frontier nodes* (the leaves of the partial decision-tree). Then, for computational simplicity, these estimates are assumed to perfect, thus licensing a problem-solver to invoke the straightforward constraint-satisfaction algorithm described above.

---

*This research was made possible by support from Heuristicrats, the National Aeronautics and Space Administration, and the Rand Corporation.

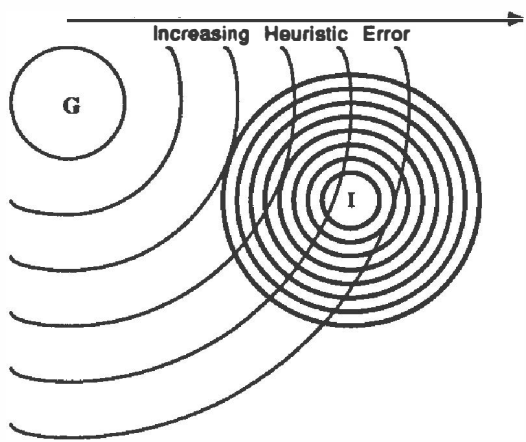

Figure 1: Search Horizon and Heuristic Error

In assuming the labels to be accurate, these traditional algorithms are liable to be fooled by error in the outcome estimates. However, it is generally assumed that this weakness of the *face-value* assumption can be compensated for by searching deeper in the tree.

The belief that error is dimished by searching deeper stems from the assumption that the heuristic error grows proportional to distance from the goal. As searching deeper causes some of the frontier nodes to approach the goal, the increased accuracy of these estimates is claimed to improve decision-quality. Thus the traditional algorithms are expected to converge to correct decisions with deeper search.

As Figure 1 suggests, this line of reasoning is flawed. As comparatively few of the nodes expanded in a search are closer to $G$ than $I$ is (exponential branching compounds the effect visible in this planar graph), most of the estimates at the frontier are fraught with error. Even under optimistic assumptions about heuristic error (e.g., bounded relative error), the likelihood of a traditional technique being misled by an erroneous estimate will increase with search depth (analytical and empirical studies can be found in [9]). Only if a search algorithm includes a sound inference mechanism for interpreting heuristic estimates can one unequivocally conclude that increasing search depth yields better decisions.

## 2 A Bayesian Approach

The Bayesian approach attempts to adjust the heuristic estimates in light of information from other nodes in the tree. Specifically, by modelling the error in the heuristic function as well as inter-node outcome constraints, one may determine, for each node, the probability of each possible outcome, conditioned on evidence provided by heuristic evaluations. In path-planning, for example, one would determine, for each node in the search graph, the probability distribution over possible distances from the nearest goal. In chess, one would determine the probability that each node leads to a win, loss or draw. Subsequently, one could take the action which maximizes expected utility[6].

To formalize this discussion, consider the search graph shown in Figure 2. From the root node, $S_0$, one must choose whether to move to $S_1$, $S_2$ or $S_3$. Let the (unknown) outcome of node $S_i$ be denoted by $O_i$. To make a rational choice (i.e., maximize expected utility) one needs the beliefs

$$P(O_1 = a), P(O_1 = b), \ldots$$
$$P(O_2 = a), P(O_2 = b), \ldots$$
$$P(O_3 = a), P(O_3 = b), \ldots$$

where $a, b, \ldots$ are the possible outcomes. Unambigously, let $P(O_i)$ denote the vector of values $(P(O_i = a), P(O_i = b), \ldots)$.

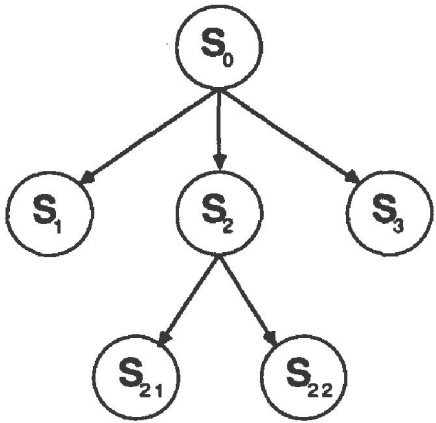

Figure 2: Search Graph

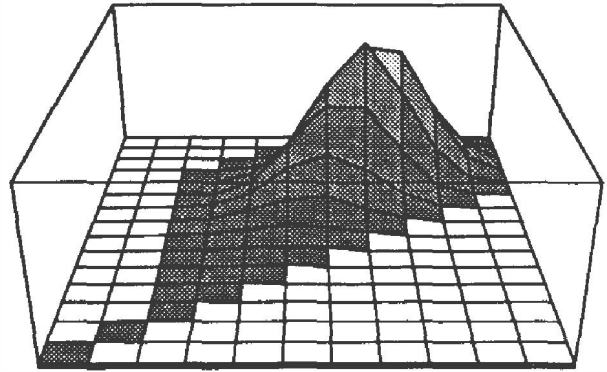

Figure 3: Probability distribution over Manhattan Distance Heuristic values (x) and Eight Puzzle outcomes (y). $P(O_i|h(S_i))$ is a slice of this distribution (a fixed x value). As the heuristic is admissible, it always underestimates the actual distance. Note that low heuristic values are perfect estimates of distance.

Once a heuristic evaluation for all the nodes in the search graph has been recorded, one must determine the three vectors

$$P(O_i|h(S_0), h(S_1), h(S_2), h(S_3), h(S_{21}), h(S_{22}))$$

where $i \in \{1, 2, 3\}$ and $h(S_i)$ is the heuristic evaluation of node $S_i$. This requires a model of the probabilistic relationships between heuristic values and outcomes, and of the inter-node outcome constraints.

## 2.1 Probabilistic Heuristic Estimates

To determine

$$P(O_i|h(S_0), h(S_1), h(S_2), h(S_3), h(S_{21}), h(S_{22}))$$

requires an association between the values returned by a heuristic evaluation function and the (unknown) outcome of the node which was evaluated. For example, if the only available evidence is the heuristic evaluation of node $S_1$, then one would need to know $P(O_1|h(S_1))$ to make a decision. In the best case, $h(S_i)$ would determine $O_i$ with certainty, and no further search would be required.

This probabilistic interpretation of a heuristic evaluation function, which we will refer to as a *probabilistic heuristic estimate* (PHE) [8], is a heuristic function $h$ together with a conditional probability distribution $P(O_i|h(S_i))$. An example is offered by Figure 3, which shows the probability distribution over Manhattan Distance heuristic values and Eight Puzzle outcomes (i.e., shortest-path lengths).

This method of interpreting heuristic evaluations allows new data to "calibrate" the PHE over time, as additional problem instances are solved, and does not require *a priori* assumptions about the accuracy of $h$. Additionally, this representation allows the principled combination of several heuristics or low-level features to create more discriminating heuristic functions.

## 2.2 Outcome Constraints

The PHE provides information about the outcomes of individual nodes. Additional knowledge is required to link these individual pieces of evidence, constrain the interpretation of each in light of the others, and yield a global interpretation to inform a decision.

In a path-planning domain, an obvious constraint is that the outcomes of adjacent nodes differ by at most the cost of the arc connecting them. For example, if node A is 17 feet from the nearest goal, and is one foot away from node B, then B's distance to the goal must be between 16 and 18 feet (see Figure 4). Such constraints allow us to integrate new information as it is uncovered, reducing the number of consistent global interpretations of the search graph. For example, on learning that node C is 19 feet away, one may compose the constraints and infer that the outcome for B is 18. This logical inference is merely a special case of probabilistic inference: learning only that C is *likely* to be 19, one should increase one's belief that B is 18.

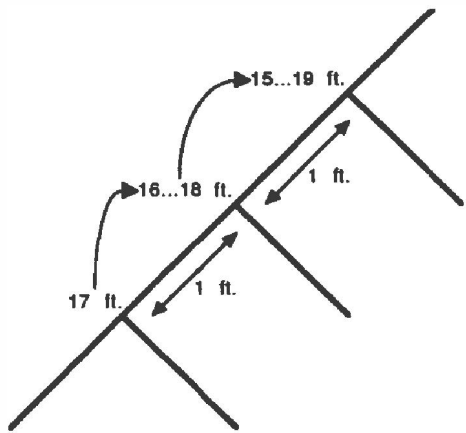

Figure 4: Path Planning Outcome Constraints

The implication for problem-solvers is that information provided by nodes deep in the tree can tightly constrain the $P(O_i)$, reducing uncertainty, and therefore enabling a more informed decision. Precisely, after evaluating a node $S_e$, one's belief in different outcomes is $P(O_i|h(S_e))$. Searching further, and evaluating $n$ *additional* nodes $\{S_{e_1}, \ldots, S_{e_n}\}$ in the subtree below $S_i$, changes one's belief to $P(O_i|h(S_e), h(S_{e_1}) \ldots, h(S_{e_n}))$.

## 3 Bayesian Inference in Search

The previous section presented the problem of inference in search, which is analogous to fitting a surface (e.g., a terrain map) to raw data. Heuristic functions act as unreliable sensors, calibrated through the PHEs, while outcome constraints embody smoothness constraints on the surface.

In this section, we discuss the mechanics of performing Bayesian inference in the search graph. As they are equivalent to those used in Bayesian Networks, we follow the notation

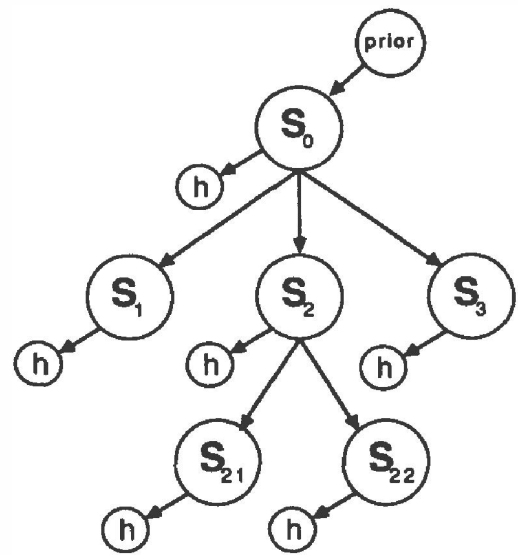

Figure 5: Search Graph as Bayesian Network

of [19].

Viewing the search graph (together with probabilistic heuristic estimates) as a Bayesian Network (Figure 5), each node $S_i$ has an associated *belief vector*, $P(O_i)$, or more properly $P(O_i|T)$, where $T$ denotes the evidence observed in the partially explored tree. Primarily, one is concerned with the belief vectors of the nodes adjacent to the root, $P(O_1), P(O_2), P(O_3)$, which ultimately will form the basis of the decision.

Consider that the heuristic function has been evaluated only at nodes $S_0$ and $S_2$. With no additional information, $P(O_2|T)$, in this case $P(O_2|h(S_2), h(S_0), O_0)$, could only be evaluated by storing and examining an enormous joint probability distribution

$$P(h(S_2), O_2, h(S_0), O_0)$$

Fortunately, this can be decomposed by the chain rule to yield

$$P(h(S_2) \mid O_2, h(S_0), O_0) P(O_2 \mid h(S_0), O_0)$$
$$P(h(S_0) \mid O_0) P(O_0)$$

and further decomposed due to conditional independence to yield

$$P(h(S_2) \mid O_2) P(O_2 \mid O_0) P(h(S_0) \mid O_0) P(O_0)$$

as the outcome constraints are assumed to specify all known dependencies.

In the general case, the decomposition of $P(O_i|T)$ yields

$$\alpha P(O_i \mid T^{+S_i}) P(h(S_i) \mid O_i) \prod_{k \in c(i)} P(T^{-S_k} \mid O_i)$$

where $T^{-S_k}$ denotes the nodes in the tree rooted at $S_k$, $T^{+S_i}$ denotes $T - T^{-S_i}$, $\alpha$ is a normalizing constant, and $c(i)$ is the set of children of $i$. Imagine the terms in this formula as being provided to the node $S_i$ via messages from its neighbors in the tree. Its parent sends it a vector of values $P(O_i \mid T^{+S_i})$, its children each provide it with a vector message $P(T^{-S_k} \mid O_i)$, and its heuristic value provides it with a vector message $P(h(S_i) \mid O_i)$. Given these incoming messages, $S_i$ can update its belief and compute the corresponding outgoing messages.

Note that in this procedure, the evaluations of internal nodes (e.g., $h(S_2)$), considered superfluous by traditional search algorithms, enable a more discerning interpretation of frontier node evaluations (this resurrects an abandoned idea from Turing's hand-simulated chess program[16]).

**Example** To illustrate the updating of beliefs based on evidence, consider the computation of the belief vector of the root node, $S_0$, by considering the messages that are sent up the tree, beginning with the leaves.

Initially, the heuristic nodes must send messages to their respective parents. If, for example, we had evaluated node $S_{21}$, with the heuristic reporting $h(S_{21})$, then the value for the corresponding variable would be fixed. The node $h_{21}$ sends to its parent, $S_{21}$, a vector message (ranging over values of $O_{21}$) indicating $P(h_{21} \mid O_{21})$. Consider that subsequently each heuristic node does the same.

At this point, a node which has received vector messages from all of its children, e.g., $S_{21}$, can compute $P(\text{evidence below } S_{21} \mid O_{21})$ or $P(T^{-S_{21}} \mid O_{21})$, by multiplying these vectors and normalizing. It can then send a vector message (ranging over values of $O_2$) to its parent indicating $P(T^{-S_{21}} \mid O_2)$, computed as $\sum_{O_{21}} P(T^{-S_{21}} \mid O_{21}) P(O_{21} \mid O_2)$. This process repeats until $S_0$ has collected messages indicating for each $O_0$, $P(\text{all evidence} \mid O_0)$, i.e., $P(T \mid O_0)$.

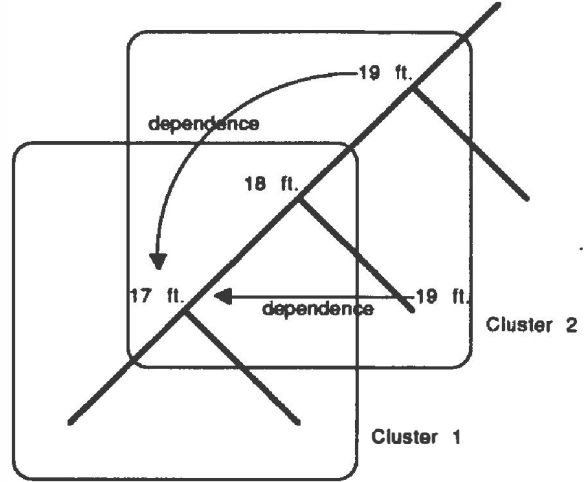

Figure 6: Source/Sink Outcome Constraints

Multiplying by the vector of values $P(O_0)$, the prior belief in the outcomes of the problem, and normalizing, yields the belief vector $P(O_0 \mid T)$ by Bayes' Rule. A similar process yields a belief vector in each node (including heuristic nodes). As $P(O_1)$, $P(O_2)$, and $P(O_3)$ are now known, a decision maximizing expected utility can be made.

To perform inference in this manner requires a constant number of vector multiplications for each node in the search graph. Each vector multiplication is linear in the number of outcomes or heuristic values (a constant over different problem instances). The space required is proportional to the depth of search, not the number of nodes. Thus, the asymptotic time and space complexity of this inference algorithm equals the lower bound over all tree-search algorithms.

### 3.1 Sophisticated Dependencies

More sophisticated dependencies may also be modelled. For example, in path-planning, there is typically a "sink/source" constraint (i.e., only a goal node can be closer to a goal than all of its neighbors (see Figure 6)).

This type of knowledge introduces additional dependencies into the search graph. The outcome of a node $S_i$ becomes dependent on the outcomes of both its neighbors and those nodes two arcs away. However, knowing the outcomes of all of these nodes renders $O_i$ independent of the outcomes of all other nodes in the graph. Similarly, in chess, a node's outcome is not only dependent on its parent, but also on its siblings, as pictured in Figure 7. Particular domains may require that more dependencies be modelled, yet there appear to be no general dependencies beyond these simple ones.

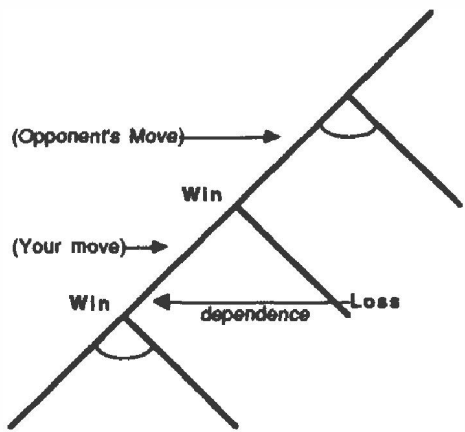

Figure 7: Game Playing Outcome Constraints

These dependencies introduce loops in the graph, which invalidates the use of the tree-based inference algorithm described above. Thus, a method is required for breaking the loops, for example, by clustering the nodes together into "macro-nodes" whose interdependencies form a tree[19] (see Figure 6). Processing such clusters increases the complexity of inference by a constant factor, as one must compute probability distributions over all combinations of outcomes in each cluster. Additionally, many heuristic functions will have the property that their values are dependent on the evaluations of neighboring nodes (e.g., consistency). These dependencies may also be modelled and clustered.

Another important though hidden dependency arises from modelling the search graph as a tree, which causes "double-counting" of evidence, as many nodes in the tree may represent the same problem state. Experimental results suggest that this may have visible effects even in a small search tree. As the problem of propagation of belief on a graph-structured Bayesian network has been shown to be NP-Hard[1], we are investigating approximation techniques for performing inference on graphs to overcome this simplifying assumption.

## 4 Empirical Tests

Minimin is a simple path-planning algorithm which consists of a full-width search to a fixed-depth, followed by a single move toward the leaf node with the minimal heuristic evaluation (this process is repeated until a goal is reached). Minimin is interesting because it parallels both Minimax and $A^*$, the classic two-player and single-agent AI algorithms (it is a one-player version of Minimax which behaves identically to time-limited $A^*$). The performance of Minimin was investigated in [5], and embedded in other planning algorithms (e.g., $RTA^*$) in [14] and [15].

We compared Minimin to BPS on the Eight Puzzle domain to study whether a sound inference procedure could produce significant performance improvements. The metric used in these experiments is the algorithm's *decision-quality* – the probability that it makes a move toward the goal – which we tabulate separately for each search horizon. As a control experiment, we also measured the decision-quality of an algorithm which applies randomly chosen operators.

For each horizon $d$, we exclude states whose distance to the goal is less than $d$ because in these cases, search would terminate at an earlier horizon. Note that this experimental condition, combined with the regular structure of the Eight Puzzle state-space, cause the decision-quality of a random algorithm to increase with search depth. Consider, for example, the limiting case of the furthest node from the goal, from which all moves are toward the goal. As is evidenced by Figure 8 (curve D), this phenomenon becomes increasingly likely as distance from the goal increases.

The decision-quality of Minimin on 10000 ran-domly selected Eight Puzzle problem instances is plotted in Figure 8, the result of our replication of the experiment of [14]. As Figure 8 (curve A) shows, Minimin appears to make better decisions as more search is performed.

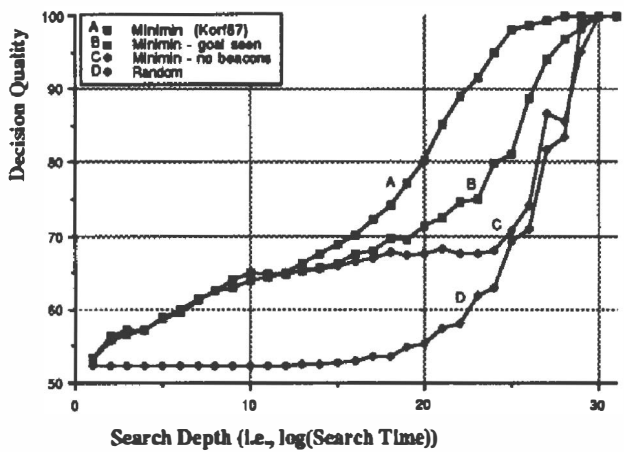

Figure 8: Decision-Quality vs. Search Depth for Minimin

However, Minimin's improvement in decision-quality is primarily due to the presence of *beacons*, or perfect information nodes, within the search horizon. An example beacon is a node for which the heuristic value perfectly estimates actual distance. As Minimin interprets heuristics at face-value, its decision-quality is artificially inflated by the increasing likelihood that beacons will appear as search horizon is increased. Unfortunately, one cannot depend on the availability of beacons in complex problems.

Hypothesizing that the results in [14] are primarily a product of unrealistic heuristic accuracy and problem simplicity, we eliminated the perfect information from the search by altering the heuristic function at the 17 beacons in the Eight Puzzle state-space (of 181440 states total). The beacons are those states where Manhattan Distance $\in \{0, 1, 2, 3\}$ – of these, 15 are within 3 moves of the goal state. These states were instead given heuristic evaluations of $MD = 4$, and the experiments repeated (curve C). In addition, an experiment was run in which the goal state provided the only beacon (curve B).

As Figure 8 (curve C) indicates, removing beacons cripples Minimin, which quickly reaches a performance plateau, in which the marginal improvement in decision-quality of further search is not worth the cost. Eventually, Minimin is little better than purely random decision-making, foreboding its performance on harder problems. Much of Minimin's performance in curve A is due to a single beacon, the goal state, as indicated by curve B. Clearly, algorithms which claim to make intelligent inferences under uncertainty cannot secretly rely on perfect information.

### 4.1 Comparing BPS and Minimin

For the purpose of comparison, we tested a simple version of BPS, which performed Bayesian inference on a full-width, fixed-depth search tree, followed by a single move to the node with the shortest expected distance to the goal. The PHE was that of Figure 3, compiled from solutions to all Eight Puzzle problem instances. BPS was tested on a random sample of 1000 problem instances – the sample problems were sufficiently difficult that BPS never had perfect information available to it. The results are shown in Figure 9 (curve A).

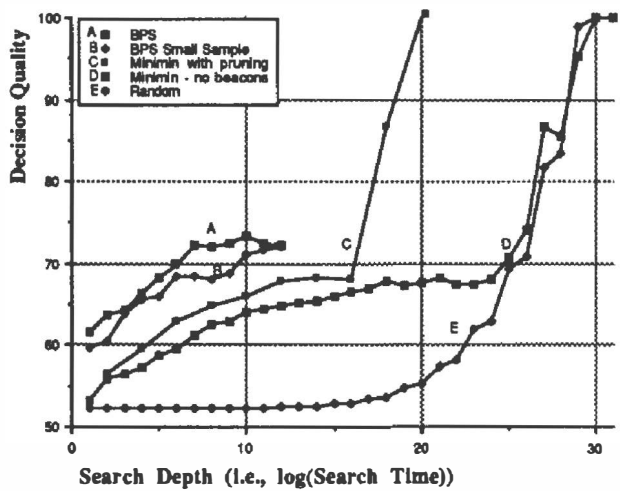

Figure 9: Decision-Quality vs. Search Depth for BPS and Minimin

BPS exceeded 70% decision-quality after less than 7 levels of lookahead (about 175 nodes), while Minimin (Figure 9, curve D) required 25

levels of lookahead (about 6 million nodes) to reach that same decision-quality. Even this is a generous comparison, as the quality of random decisions from nodes at least 25 moves from the goal is itself nearly 69%.

The PHE used in the above test is somewhat unrealistic, as it is derived from exploration of the complete state-space. Also shown in Figure 9 (curve B) are the performance results of BPS when equipped with a PHE interpolated from a sample of 1000 random Eight Puzzle problem instances, together with the 500 nodes nearest to the goal. The smaller sample slightly degrades the decision-quality of BPS, however it still outperforms Minimin by orders of magnitude (requiring approximately 750 nodes to achieve 70% decision-quality, in contrast to 6 million nodes for Minimin).

### 4.2 Control of Inference

As has been shown in [15], however, a pruning mechanism exists for Minimin, which allows it to search approximately 50% deeper on the Eight Puzzle than regular Minimin, in a given amount of time. Effectively, this shifts the Minimin curve (in Figure 9) one-third closer (from curve D to curve C) to the BPS curve (curve A). Despite the computational advantage of Minimin's pruning, its decision-quality is only marginally improved. It is important to note that the results described above stem only from the power of BPS' probabilistic inference mechanism, to which one could add a selective evidence-gathering mechanism, the analogue of the pruning strategies used by many search algorithms.

Most branch-and-bound techniques (e.g., $A^*$ and $\alpha$-$\beta$-pruning) which assume the face-value interpretation of heuristics, prune by ignoring portions of the search tree which can be *proven* to have no relevance to the ultimate decision. Under conditions of uncertainty, however, few of the common pruning techniques can provide such guarantees. The explicit representation of probability within BPS affords the problem-solver the opportunity to direct search using standard decision-theoretic mechanisms for ordering node expansion (i.e., value of information[12][13][21]).

In principle, a search algorithm, just as it chooses the moves to make by considering the likelihood of different outcomes of an action, should choose low-level actions (e.g., performing heuristic evaluations, expanding nodes) with the same considerations in mind. To do so, it must be able to anticipate the immediate results of each action (what values can a heuristic assume? how will that affect current beliefs?), and relate those results to long-term goals (will this information improve decision-quality?). Initial results in this direction are reported in[7]. Similar research, on decision-theoretic control, is being pursued by a number of researchers, as applied to resource allocation in planning [2][3], control of probabilistic inference[11], and control of traditional search algorithms[5][23][24], having been suggested over twenty years ago in [4].

## 5 Conclusion

In this paper, we have shown how Bayesian probability may be applied to the straightforward inferences required in search. In contrast, most search algorithms choose the default calculus of ignoring uncertainty in heuristic information. Others, which do not adhere to the face-value principle, represent alternative inference calculi for reasoning about uncertainty [10].

Further, we have demonstrated how sound uncertainty management enables decisions superior to those of an accepted, *ad hoc* technique, and does so with significantly less computational effort. Together, these indicate that a fundamental method underlying many AI systems, heuristic search, can be profitably viewed as a problem of inference from uncertain evidence.

Thus, we conjecture that in addition to the often-noted trade-off between "knowledge" and "search" in artificial intelligence, there is a third, and overlooked axis, "inference". In heuristic search, for example, "knowledge" refers to the accuracy of the heuristic evaluation function, and "search", the number of nodes expanded (see ch. 6 of [18] for quan-titative analyses in the context of the $A^*$ algorithm, and [22] for a general discussion of the trade-off in AI systems). The third axis, "inference", refers to the interpretation of the knowledge accumulated from search – the combination of evidence. Our experimental results, holding knowledge constant, suggests that a little inference is worth a lot of search.

## Acknowledgements

We thank Jack Breese, Peter Cheeseman, Eric Horvitz, Judea Pearl and Stuart Russell for many stimulating discussions concerning this research, and Mark Boddy, Gerhard Holt, Beth Mayer, Barney Pell, Eric Wefald and Mike Wellman for their comments on earlier drafts of this paper.